\documentclass[runningheads]{llncs}

\usepackage{hyperref}
\usepackage[utf8]{inputenc}
\usepackage[T1]{fontenc}
\usepackage{url}
\usepackage{amsmath}
\usepackage{amssymb}
\usepackage{mathtools}
\usepackage{graphicx}
\usepackage{paralist}
\usepackage{textcomp}
\usepackage{gensymb}
\usepackage{nicefrac}
\usepackage[dvipsnames]{xcolor}

\begin{document}

\frontmatter          

\pagestyle{headings}  
\addtocmark{}
\mainmatter              

\title{Graph of brain structures grading for early detection of Alzheimer's disease}

\titlerunning{Graph of brain structures grading for early detection of Alzheimer's disease}

\author{%
Kilian Hett \inst{1,2} \and %
Vinh-Thong Ta\inst{1,2,3} \and %
Jos\'e V. Manj\'on\inst{4} \and 
Pierrick Coupé\inst{1,2} and the 
Alzheimer's Disease Neuroimaging Initiative\protect\footnote[1]{\scriptsize{Data used in preparation of this article were obtained from the Alzheimer's Disease Neuroimaging Initiative (ADNI) 
database (adni.loni.usc.edu).  As such, the investigators within the ADNI contributed to the design and implementation of ADNI and/or provided data but
did not participate in analysis or writing of this report.  A complete listing of ADNI investigators can be found at: 
\url{http://adni.loni.usc.edu/wp-content/uploads/how_to_apply/ADNI_Acknowledgement_List.pdf}.}}
}

\authorrunning{K. Hett et al.}

\institute{%
Univ. Bordeaux, LaBRI, UMR 5800, PICTURA, F-33400 Talence, France\\
\and
CNRS, LaBRI, UMR 5800, PICTURA, F-33400 Talence, France\\
\and
Bordeaux INP, LaBRI, UMR 5800, PICTURA, F-33600 Pessac, France\\
\and
Universitat Polit\`ecnia de Val\`encia, ITACA, 46022 Valencia, Spain
}

\maketitle              

\begin{abstract}
Alzheimer's disease is the most common dementia leading to an irreversible neurodegenerative process. 
To date, subject revealed advanced brain structural alterations when the diagnosis is established. 
Therefore, an earlier diagnosis of this dementia is crucial although it is a challenging task. 
Recently, many studies have proposed biomarkers to perform early detection of Alzheimer's disease. 
Some of them have proposed methods based on inter-subject similarity 
while other approaches have investigated framework using intra-subject variability.  
In this work, we propose a novel framework combining both approaches
within an efficient graph of brain structures grading. Subsequently, we demonstrate the competitive performance 
of the proposed method compared to state-of-the-art methods.

\end{abstract}

\section{Introduction}

Alzheimer's disease (AD) is the most common dementia leading to a neurodegenerative process causing mental dysfunctions. According to the 
world health organization, the number of patients having AD will double in 20 years. Neuroimaging studies performed on AD subjects revealed 
that brain structural alterations are advanced when diagnosis is established. Indeed, the clinical symptoms of AD is preceded by brain changes 
that stress the need to develop new biomarkers to detect the first stages of the disease. 
The development of such biomarkers can make easier the design of clinical trials and 
therefore accelerate the development of new therapies.

Over the past decades, the improvement of magnetic resonance imaging (MRI) has led to the development of new imaging biomarkers \cite{bron2015standardized}.
Many works developed biomarkers based on inter-subject similarities 
to detect anatomical alterations by using group-based comparison (\emph{e.g.}, patients vs. normal controls). 
Some of them are based on regions of interest (ROI) to capture brain structural alterations at a large scale of analysis.
The alterations of specific structures such as the cerebral cortex and hippocampus (HIPP) are usually captured with volume, shape, or cortical thickness (CT) measurements \cite{wolz2011multi}.
Other approaches proposed to study the inter-subject similarity between individuals from the same group at a voxel scale. 
Such methods commonly use voxel-based morphometry (VBM). VBM-based studies showed that the
medial temporal lobe (MTL) is a key area to detect the first manifestations of AD \cite{wolz2011multi}. 
Recently, more advanced methods have been designed to improve computer-aided diagnosis \cite{bron2015standardized}. 
Among them, patch-based grading (PBG) framework  \cite{coupe2012scoring} proposed to better analyze inter-subject similarities.
PBG demonstrated state-of-the-art results for AD diagnosis and prognosis
\cite{coupe2012scoring,hett2017adaptive,tong2017novel}.

Beside inter-subject similarity approaches, other methods proposed to capture the correlation of brain structures alterations within subjects.
Indeed, although similarity-based biomarkers provide helpful tools to detect the first signs of AD, the structural alterations leading 
to cognitive decline are not homogeneous within a given subject. 
Such biomarkers assumed that the structural changes caused by the disease may not occur at isolated areas but in several inter-related regions.
Therefore, intra-subject variability features provide relevant information.
Some methods proposed to capture the relationship of spread cortical atrophy 
with a network-based framework \cite{wee2013prediction}. 
Other approaches estimate inter-regional correlation of brain tissues volumes \cite{zhou2011hierarchical}. 
Recently, convolutional neural network (CNN) have been used to capture relationship between anatomical structures volumes \cite{suk2017deep}.
Finally, some works showed that patch-based strategy can be used to model intra-subject brain alteration \cite{liu2014hierarchical,tong2014multiple}.

The main contribution of this work is the development of a novel representation based on a graph of brain structures grading (GBSG) combining inter-subject pattern similarity and intra-subject variability features 
to better capture AD signature. 
First, inter-subject similarities are captured using patch-based grading framework applied over the entire brain. 
Second, intra-subject variabilities are modeled by a graph representation.
In our experiments, we compare the performance of intra-subject variability features (\emph{i.e.}, the edges of our graph) with inter-subject pattern similarity features 
(\emph{i.e.}, the vertices). 
Moreover, we demonstrate the capability of intra-subject variability features to early detect AD and show that the combination of both features
improves AD prognosis. 
Finally, we present competitive results of our new method compared to state-of-the-art approaches.

\section{Materials and Methods}
\subsection{Dataset}
Data used in this work were obtained from the Alzheimer's Disease Neuroimaging Initiative (ADNI) dataset\footnote[1]{\url{http://adni.loni.ucla.edu}}. 
We use all the baseline T1-weighted (T1w) MRI of the ADNI1 phase. 
This dataset includes AD patients, subjects with mild cognitive impairment (MCI) and cognitive normal (CN) subjects (see Table~\ref{tab:dataset}).
MCI is a presymptomatic phase of AD composed of subjects who have abnormal memory dysfunctions. In our experiments we consider two groups of MCI. 
The first group is composed of patients having stable MCI (sMCI) and the second one is composed with patients having 
MCI symptoms at the baseline and converted to AD into the following 36 months. This group is named progressive MCI (pMCI).
\begin{table}
\centering
\caption{Description of the ADNI dataset used in this work.}
\begin{tabular}{@{\hspace{0.4cm}} l @{\hspace{0.4cm}} c @{\hspace{0.4cm}} c @{\hspace{0.4cm}} c @{\hspace{0.4cm}} c @{\hspace{0.4cm}}} \hline
Characteristic /  Group & CN & sMCI & pMCI & AD \\ \hline
Number of subjects 	& 228 & 100 & 164 & 191 \\
Ages (years) 		& $75.8 \pm 5.0$ & $75.3 \pm 7.2$ & $74.2 \pm 6.64$ & $75.26 \pm 7.4$ \\
Sex (M/F) 		& $118/110$ & $66/34$ & $97/67$ &  $100/91$ \\ 
MMSE 			& $29.05 \pm 0.9$ & $27.1 \pm 2.5$ & $26.3 \pm 2.0$ & $22.8 \pm 2.9$ \\ \hline
\end{tabular}
\label{tab:dataset}
\end{table}

\subsection{Preprocessing}
The data are preprocessed using the following steps: (1) denoising using a spatially adaptive non-local means filter \cite{manjon2010adaptive}, 
(2) inhomogeneity correction using N4 method \cite{tustison2010n4itk}, (3) low-dimensional non-linear registration to MNI152 space using 
ANTS software \cite{avants2011reproducible}, (4) intensity standardization, (5) segmentation using a non-local label fusion \cite{giraud2016optimized} 
and (6) systematic error corrections \cite{wang2011learning}.
The patch-based multi-template segmentation was performed using 35 images manually labeled  by Neuromorphometrics, Inc. \footnote{\url{http://Neuromorphometrics.com}} 
using the brain-COLOR labeling protocol composed of 134 structures.

\subsection{Computation of patch-based grading biomarkers}

To capture alterations caused by AD, we use the recently developed patch-based grading framework \cite{coupe2012scoring}. 
PBG framework provides at each voxel a grade between $-1$ and $1$ related to the alteration severity.
The grading value $g$ at $x_i$ is defined as:
\begin{equation}
g_{x_i} = \frac{\sum_{t_{j} \in K_{i}} w(P_{x_i},P_{t_{j}}) p_t}{\sum_{t_{j} \in K_{i}} w(P_{x_i},P_{t_j})}, \label{eq:grade}
\end{equation} 
where $P_{x_i}$ and $P_{t_{j}}$ represent the patches surrounding the voxel $i$ of the test subject image $x$ 
and 
the voxel $j$ of the template image $t$, respectively. 
The template $t$ comes from a training library composed of CN subjects and AD patients. 
$p_t$ is the pathological status set to $-1$ for patches extracted from AD patients and to $1$ 
for those extracted from CN subjects. 
$K_i$ is a set of the most similar $P_{t_j}$ patches to $P_{x_i}$ found in the training library. 
The anatomical similarity between the test subject $x$ and the training library is estimated by a weight function
$w(P_{x_{i}},P_{t_{j}}) = \exp (-  ||P_{x_{i}} - P_{t_{j}}||^2_2 / ( h^2 +\epsilon ) )$, 
where $h=\min_{t_{j}} ||P_{x_i} - P_{t_{j}}||^2_2$ and $\epsilon \to 0$.

\subsection{Graph construction}

\begin{figure*}[!ht] 
\centering
\includegraphics[width=1\linewidth]{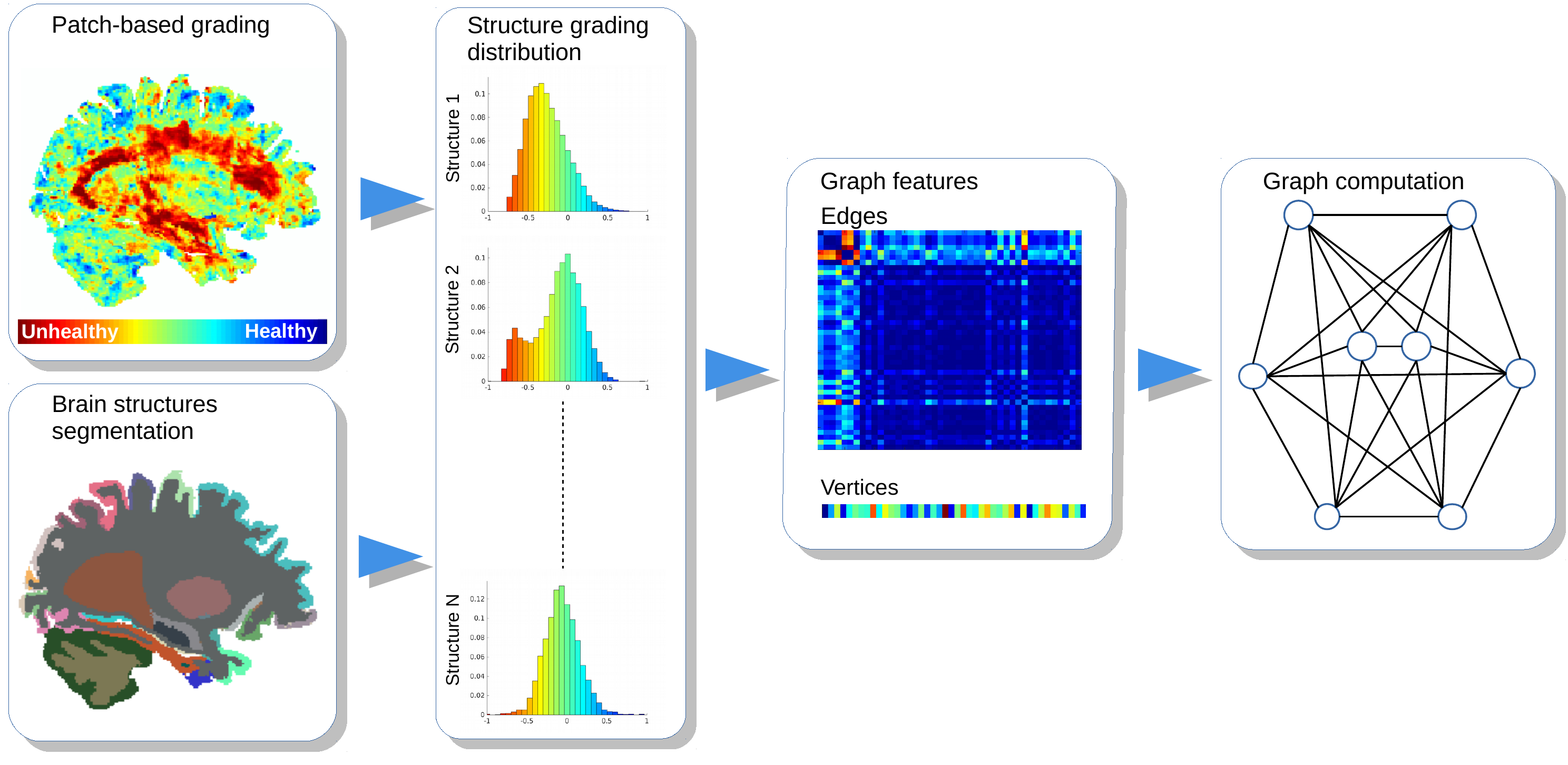}
\caption{Illustration of the graph construction method. From left to right, for each segmented structure an estimation of the 
density probability of PBG values are computed. 
Then, histograms are used to built our graph of brain structure grading. Thus, 134 histograms representing each segmented brain structures are estimated. 
Edges are the distances between structure grading distribution while vertices are the mean grading value for a given structure 
(see text for more details).  
}
\label{fig:segmentation}
\end{figure*}

In our GBSG method, the grading process is carried out over the entire brain.
Afterwards, the corresponding segmentation is used to fuse grading values
and to built our graph (see Fig. \ref{fig:segmentation}). 
We define an undirected graph $G=(V, E, \Gamma, \omega)$, where $V=\{v_1,...,v_N\}$ is the set of vertices 
for the $N$ considered brain structures and
$E=V \times V$ is the set of edges.
In our work, the vertices are the mean of the grading values for a given structure while the edges are based on 
grading distributions distances between two structures (see Fig.~\ref{fig:segmentation}).

To this end, the probability distributions of PBG values (see Eq.~\ref{eq:grade}) 
are estimated with a histogram $H_v$ for each structure $v$.
The number of bins 
is computed with the Sturge's rule \cite{sturges1926choice}. 
For each vertex we assign a function $\Gamma:V\rightarrow \mathbb{R}$ defined as $\Gamma(v) = \mu_{H_v}$, 
where $\mu_{H_v}$ is the mean of $H_v$.
For each edge we assign a weight given by the function $\omega:E\rightarrow \mathbb{R}$ defined as
$\omega(v_i,v_j) = \exp ( -d(H_{v_i},H_{v_j})^2 / \sigma^2  )$ 
where $d$ is the Wasserstein distance with $L_1$ norm \cite{rubner2000earth} that showed best performance during our experiments.

Graph representation of structure grading provides high-dimensional features (see Fig.~\ref{fig:segmentation}). 
In this work we used the Elastic Net regression (EN) method that provides a sparse representation of the most discriminative edges and vertices, 
and thus enables to reduce the feature dimensionality by capturing the key structures and the key relationships between the different brain structures 
(see Fig.~\ref{fig:workflow}).
Thus, after normalization, a concatenation of the two feature vectors is given as input of EN feature selection method.

\subsection{Details of implementation}
\begin{figure*}[!ht]
\centering
\includegraphics[width=1\linewidth]{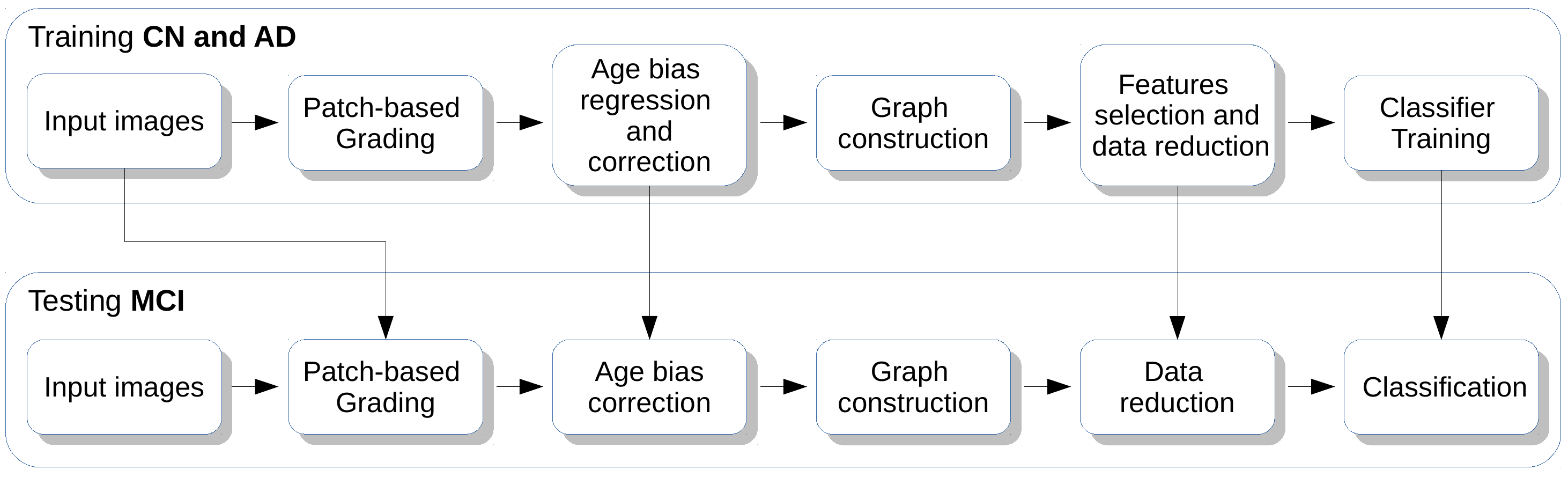}
\caption{Scheme of the proposed GBSG pipeline. PBG is computed using CN and AD training groups. 
CN group is also used to correct the bias related to age.
This estimation is applied then to AD and MCI subjects. Afterwards, the graph is constructed, 
the feature selection is trained on CN and AD and is applied to CN, AD and 
MCI. Finally the classifier is trained with CN and AD.
}
\label{fig:workflow}
\end{figure*}

PBG was computed with a patch-match method \cite{giraud2016optimized}.
We used the parameters proposed in \cite{hett2017adaptive} for patch size and size of $K_i$.
This results in a whole brain grading in about 10 seconds. 
Age effect is corrected using linear regression estimated on CN population.
The EN method is computed with SLEP package \cite{liu2009slep}. 
Two classifiers were used to validate our method – a support vector machine (SVM)~\footnote{\url{http://www.csie.ntu.edu.tw/~cjlin/libsvm}}.
with a linear kernel and a random forest (RF)~\footnote{\url{http://code.google.com/p/randomforest-matlab}}. 
The linear SVM has a soft margin parameter, 
which was optimized in a range of $2^i$, with $i=\{-10,9,...,10\}$. 
All features were normalized using z-score. 
In our experiments, we performed sMCI versus pMCI classification.
The EN features selection and the classifiers were trained with CN and AD (see Fig.~\ref{fig:workflow}).
As shown in \cite{tong2017novel}, the use of CN and AD to train the feature selection method and the classifier enables to better discriminate sMCI and pMCI subjects. 
Moreover, it also enables to get the results without cross-validation step and more importantly to limit bias and over-fitting problem.
Thus, only one run was performed for the SVM and 30 runs was performed to capture the inner variability of RF.
The mean accuracy (ACC), sensibility (SEN), and specificity (SPE) over these 30 iterations are provided as results (see Table \ref{tab:results}).


\section{Results and Discussions}
\begin{table}
\centering
\caption{Classification of sMCI versus pMCI. Results 
obtained by inter-subject similarity features (\emph{i.e.}, vertices), intra-subject variability features (\emph{i.e.}, edges) 
and a combination of both. The original PBG applied on HIPP \cite{coupe2012scoring} is used as baseline.
Results show that GBSG edge features improve the accuracy and the sensibility as compared to HIPP PBG and GBSG vertices features. 
Finally, GSBG provides the best results.}
\begin{tabular}{@{\hspace{0.8cm}} l @{\hspace{0.8cm}} c @{\hspace{0.8cm}} c @{\hspace{0.8cm}} c @{\hspace{0.8cm}} c @{\hspace{0.8cm}} c @{\hspace{0.8cm}}} \hline
Methods & Classifier & ACC & SEN & SPE \\ \hline
HIPP PBG & SVM & 71.5\%  & 72.5\% & 70.0\% \\ 
GBSG Vertices & SVM  & 71.9\% & 71.95\% & \textbf{72.0\%} \\ 
       & RF  & 70.1\% & 69.6\% & 71.1\% \\
GBSG Edges & SVM & 74.6\%  & 81.7\% & 63.0\% \\ 
& RF  & 73.8\%  & 81.3\% & 61.6\% \\
GBSG & SVM & 75.8\%   & 82.3\% & 65.0\% \\
 & RF & \textbf{76.5\%} & \textbf{81.7\%} & 68.0\% \\ \hline
\end{tabular}
\label{tab:results}
\end{table}

To investigate the results of our new GBSG method combining inter-subject pattern similarity features (\emph{i.e.}, vertices) 
and intra-subject variability features (\emph{i.e.}, the edges) several experiments were performed (see Table~\ref{tab:results}). 
The original hippocampal grading (HIPP PBG) is used as baseline \cite{coupe2012scoring}. 

First, we estimated the classification performances obtained by each feature separately using SVM.  
Compared to HIPP PBG, vertices showed an improvement of the specificity while the accuracy and the sensibility did not change. 
Therefore, additional structures selected by EN did not improve results compared to use HIPP only. 
On the other hand, the edges feature improved the accuracy and the sensibility but was less specific compared to HIPP PBG and vertices.
These results indicate that relevant information is encoded within GBSG edges.

Second, we evaluated the performance of combining vertex and edge features. 
GBSG provided the best results in terms of accuracy and sensibility. 
Moreover it improved the specificity compared to the intra-subject variability features. 
Finally, we compared SVM and RF classifiers to study the stability of our framework. The results obtained with both classifier showed the same tendency.
The RF provided the best results with 76.5\% of accuracy.

These results obtained with two different classification methods demonstrate the complementarity
of inter-subject similarity and intra-subject variability features. Indeed,
both information – level of structure degradations and global 
pattern of key structure modifications – are relevant.

Afterwards, we compared our GBSG method using RF classifier with state-of-the-art methods on similar ADNI1 datasets. 
First, we included methods modeling inter-subject variability based on 
PBG within HIPP \cite{coupe2012scoring}, VBM \cite{moradi2015machine} and an advanced PBG (aPBG) estimated over the entire brain \cite{tong2017novel}. 
Second, we included methods capturing intra-subject variability based on last deep learning framework \cite{suk2017deep}, 
multiple instance learning (MIL) \cite{tong2014multiple} and
integrative network of cortical thickness abnormality (ICT) \cite{wee2013prediction}. 
We applied our GBSG on two definitions of sMCI/pMCI populations as defined in \cite{moradi2015machine} and \cite{tong2017novel} 
to perform a fair comparison. Results of this comparison are presented in Table~\ref{tab:comparisons}.
This comparison shows that best methods based on intra-subject or inter-subject obtained similar accuracy around 75\% while our GBSG combining both reached 
76.5\% of accuracy.  
Compared to VBM on the same dataset \cite{moradi2015machine}, our GBSG improved accuracy by 1.8 percent point. 
However, compared to aPBG \cite{tong2017novel} GBSG provided similar results on the same dataset. 
Finally, compared to the CNN-based method proposed in \cite{suk2017deep} our method obtained competitive performances. 
These results highlight the efficiency of combining intra-subject and inter-subject features.

\begin{table}
 \centering
 \caption{Comparison of the proposed method with state-of-the-art approaches. 
 These results show the competitive performance of our new GBSG method that obtains the best accuracy on both definitions of sMCI/pMCI populations.}
 \begin{tabular}{@{\hspace{0.4cm}} l @{\hspace{0.4cm}} c  @{\hspace{0.4cm}} c @{\hspace{0.4cm}} c @{\hspace{0.4cm}} c @{\hspace{0.4cm}} c @{\hspace{0.4cm}} c @{\hspace{0.4cm}} c @{\hspace{0.4cm}} c @{\hspace{0.4cm}}} \hline
 Method 			& sMCI/pMCI 	& Area 		& Feature & ACC & SEN & SPE \\ \hline
 PBG \cite{coupe2012scoring} 	& 238/167	& HIPP 		& Inter & 71.0\% & 70.0\% &  71.0\%  \\
 VBM \cite{moradi2015machine} 	& 100/164 	& Brain 	& Inter & 74.7\%  & {88.8\%} & 51.59\% \\ 
 aPBG \cite{tong2017novel} 	& 129/171 	& Brain 	& Inter & 75.0\% & - & - \\  \hline
 ICT \cite{wee2013prediction} 	& 111/89  	& Cortex 	& Intra  & 75.0\%  & 63.5\% & {84.4\%} \\
 MIL \cite{tong2014multiple} 	& 238/167	& MTL 		& Intra & 72.0\% & 69.0\% & 74.0\% \\
 CNN \cite{suk2017deep}   	& 226/167	& GM 		& Intra   & 74.8\% & 70.9\% & 78.8\% \\  \hline 
 \textbf{GBSG} 			& 129/171 	& Brain 	& Inter + Intra & \textbf{75.2\%} & 80.0\% & 68.7\%  \\
				& 100/164 	& Brain 	& Inter + Intra  & \textbf{76.5\%} & 81.7\% & 68.0\% \\ \hline
 \end{tabular}
 \label{tab:comparisons}
\end{table}

\section{Conclusions}
In this paper, we proposed a novel framework based on a promising graph of brain structures grading. Our new method combines inter-subject pattern similarities 
and intra-subject variabilities to better detect AD alterations. 
The pattern similarity is estimated with a patch-based grading strategy, while the intra-subject variability between structures grading is based on graph modeling.
Our experiments showed the complementarity of both information.
Finally, we demonstrated that our method obtains competitive performance compared to the most advanced methods.

\subsubsection{Acknowledgement}
This study has been carried out with financial support from the
French State, managed by the French National Research Agency (ANR)
in the frame of the Investments for the 
future Program IdEx Bordeaux 
(HL-MRI ANR-10-IDEX-03-02), Cluster of excellence CPU and TRAIL (HR-DTI ANR-10-LABX-57).

\bibliographystyle{plain}

\begin{thebibliography}{10}

\bibitem{avants2011reproducible}
Avants et~al.
\newblock A reproducible evaluation of \textsc{ANT}s similarity metric
  performance in brain image registration.
\newblock {\em Neuroimage}, 54(3):2033--2044, 2011.

\bibitem{bron2015standardized}
Bron et~al.
\newblock Standardized evaluation of algorithms for computer-aided diagnosis of
  dementia based on structural \textsc{MRI}: The \textsc{CADD}ementia
  challenge.
\newblock {\em NeuroImage}, 111:562--579, 2015.

\bibitem{coupe2012scoring}
Coup{\'e} et~al.
\newblock Scoring by nonlocal image patch estimator for early detection of
  \textsc{A}lzheimer's disease.
\newblock {\em NeuroImage: clinical}, 1(1):141--152, 2012.

\bibitem{giraud2016optimized}
Giraud et~al.
\newblock An optimized patchmatch for multi-scale and multi-feature label
  fusion.
\newblock {\em NeuroImage}, 124:770--782, 2016.

\bibitem{hett2017adaptive}
Hett et~al.
\newblock Adaptive fusion of texture-based grading: Application to
  Alzheimer’s disease detection.
\newblock In {\em International Workshop on Patch-based Techniques in Medical
  Imaging}, pages 82--89. Springer, 2017.

\bibitem{liu2009slep}
Liu et~al.
\newblock Slep: Sparse learning with efficient projections.
\newblock {\em Arizona State University}, 6(491):7, 2009.

\bibitem{liu2014hierarchical}
Liu et~al.
\newblock Hierarchical fusion of features and classifier decisions for
  Alzheimer's disease diagnosis.
\newblock {\em Human brain mapping}, 35(4):1305--1319, 2014.

\bibitem{manjon2010adaptive}
Manj{\'o}n et~al.
\newblock Adaptive non-local means denoising of \textsc{MR} images with
  spatially varying noise levels.
\newblock {\em Journal of Magnetic Resonance Imaging}, 31(1):192--203, 2010.

\bibitem{moradi2015machine}
Moradi et~al.
\newblock Machine learning framework for early MRI-based Alzheimer's conversion
  prediction in MCI subjects.
\newblock {\em Neuroimage}, 104:398--412, 2015.

\bibitem{rubner2000earth}
Rubner et~al.
\newblock The earth mover's distance as a metric for image retrieval.
\newblock {\em International journal of computer vision}, 40(2):99--121, 2000.

\bibitem{suk2017deep}
Suk et~al.
\newblock Deep ensemble learning of sparse regression models for brain disease
  diagnosis.
\newblock {\em Medical Image Analysis}, 37:101--113, 2017.

\bibitem{tong2014multiple}
Tong et~al.
\newblock Multiple instance learning for classification of dementia in brain
  \textsc{MRI}.
\newblock {\em Medical Image Analysis}, 18(5):808--818, 2014.

\bibitem{tong2017novel}
Tong et~al.
\newblock A novel grading biomarker for the prediction of conversion from mild
  cognitive impairment to \textsc{A}lzheimer's disease.
\newblock {\em IEEE Transactions on Biomedical Engineering}, 64(1):155--165,
  2017.

\bibitem{tustison2010n4itk}
Tustison et~al.
\newblock N4\textsc{ITK}: improved \textsc{N}3 bias correction.
\newblock {\em IEEE Transactions on Medical Imaging}, 29(6):1310--1320, 2010.

\bibitem{wang2011learning}
Wang et~al.
\newblock A learning-based wrapper method to correct systematic errors in
  automatic image segmentation: consistently improved performance in
  hippocampus, cortex and brain segmentation.
\newblock {\em NeuroImage}, 55(3):968--985, 2011.

\bibitem{wee2013prediction}
Wee et~al.
\newblock Prediction of Alzheimer's disease and mild cognitive impairment using
  cortical morphological patterns.
\newblock {\em Human brain mapping}, 34(12):3411--3425, 2013.

\bibitem{wolz2011multi}
Wolz et~al.
\newblock Multi-method analysis of \textsc{MRI} images in early diagnostics of
  \textsc{A}lzheimer's disease.
\newblock {\em PloS one}, 6(10):e25446, 2011.

\bibitem{zhou2011hierarchical}
Zhou et~al.
\newblock Hierarchical anatomical brain networks for MCI prediction: revisiting
  volumetric measures.
\newblock {\em PloS one}, 6(7):e21935, 2011.

\bibitem{sturges1926choice}
Sturges.
\newblock The choice of a class interval.
\newblock {\em Journal of the american statistical association},
  21(153):65--66, 1926.

\end{thebibliography}

\end{document}